\documentclass[10pt]{article} 

\usepackage[accepted]{rlj} 

\usepackage{amssymb}            
\usepackage{mathtools}          
\usepackage{mathrsfs}           
\usepackage{graphicx}           
\usepackage{subcaption}         
\usepackage[space]{grffile}     
\usepackage{url}                
\usepackage{lipsum}             

\title{Short-Term-to-Long-Term Memory Transfer for Knowledge Graphs under Partial Observability}

\setrunningtitle{Short-Term-to-Long-Term Memory Transfer for Knowledge Graphs under Partial Observability}

\author{Taewoon Kim\textsuperscript{1,2}, Vincent Fran\c{c}ois-Lavet\textsuperscript{2}, Michael Cochez\textsuperscript{3}}

\emails{taewoon@humem.ai, vincent.francoislavet@vu.nl, michael.cochez@abo.fi}

\affiliations{
$^{1}$\textbf{HumemAI}\\
$^{2}$\textbf{Vrije Universiteit Amsterdam}\\
$^{3}$\textbf{ELLIS Institute Finland \& Abo Akademi University}
}

\contribution{ We formalize short-term-to-long-term memory transfer as a neuro-symbolic
reinforcement learning problem in partially observable knowledge-graph memory, where
each observed fact receives an explicit keep/drop decision before long-term insertion. }
{ The formulation isolates transfer decisions in the RoomKG benchmark while keeping question
answering (QA), exploration, and eviction policies fixed to avoid policy-interaction
confounds, so observed differences can be attributed to memory transfer. }

\contribution{ We introduce a per-item Q-learning formulation for variable-cardinality
short-term memory, using shared parameters to produce independent keep/drop Q-values as
the number of candidate facts changes over time. } { Methodologically, this contributes
a practical value-based training recipe for variable-cardinality memory transfer
decisions (temporal-difference (TD) updates over matched items when short-term
observation counts differ across consecutive steps, with replay-driven stochastic
pairing), complementing latent-memory RL approaches such as Deep Recurrent Q-Network
(DRQN) and external-memory models
\citep{hausknecht2017deeprecurrentqlearningpartially,graves2014neuralturingmachines,Graves2016HybridCU}.
}

\contribution{ We provide controlled empirical and behavioral evidence on the RoomKG benchmark at a
long-term memory capacity of 128: learned transfer decisions improve over symbolic and
neural baselines, including symbolic baselines with temporal annotations,
additional symbolic transfer baselines, and history-based RL baselines from prior RoomKG
work \citep{kim2026temporalknowledgegraphmemorypartially}. } { Within the learned
family, graph neural network (GNN)-based ablations show the strongest variant in this
regime is a lightweight local policy that uses short-term input only, and step-level
keep/drop analysis supports interpretable selection of task-relevant facts. }


\summary{We study short-term-to-long-term memory transfer in reinforcement learning
under partial observability with symbolic knowledge-graph memory. Our method treats each
observed fact as a keep/drop decision and learns these decisions with a per-item
Q-learning design that handles variable-sized short-term memory through shared
parameters and matched temporal-difference (TD) updates. On the RoomKG benchmark at a long-term memory
capacity of 128, learned transfer decisions outperform symbolic and neural baselines,
including symbolic baselines with temporal annotations and history-based RL baselines.
Graph neural network (GNN) ablations show that a lightweight local policy with
short-term-only input is strongest in this regime. Behavioral analysis shows that the
policy keeps navigation- and query-relevant facts while discarding lower-value candidate
facts, supporting explicit and interpretable memory decisions.}

\begin{document}

\makeCover  
\maketitle  

\begin{abstract}
Reinforcement learning under partial observability requires deciding what information to
retain, yet most memory-based approaches do not explicitly model short-term-to-long-term
transfer of symbolic observations. We study this transfer process in a temporal
knowledge-graph memory setting and cast it as a neuro-symbolic value-based decision
problem: for each observed triple, the agent chooses whether to keep or drop it before
long-term insertion. To handle variable-sized short-term buffers, we use a per-item
Q-learning design with shared parameters and a practical temporal-difference update over
matched items across consecutive steps. On the RoomKG benchmark at long-term memory capacity 128,
learned transfer decisions outperform symbolic and neural baselines, including symbolic
baselines with temporal annotations and history-based LSTM/Transformer baselines. Across transfer-policy
ablations, a lightweight local short-term-only variant performs best, and step-level
behavior shows that the policy keeps navigation- and query-relevant facts while
discarding lower-value candidate facts, supporting explicit and interpretable memory
decisions under memory constraints.
\end{abstract}


\section{Introduction}
\label{sec:introduction}

Humans routinely act under partial observability, yet still make effective decisions by
selectively moving information from immediate experience into longer-term memory. This
memory-centric perspective motivates reinforcement learning settings where performance
depends not only on action selection, but also on what information is retained.

In reinforcement learning under partial observability, many successful methods use recurrent or external-memory
architectures to learn latent states
\citep{hausknecht2017deeprecurrentqlearningpartially,graves2014neuralturingmachines,Graves2016HybridCU,pritzel2017neuralepisodiccontrol}.
While effective, these approaches usually do not model an explicit neuro-symbolic transfer
decision: when an observation arrives, which facts should be transferred to long-term
memory and which should be dropped.

We study this question in a temporal knowledge-graph memory setting \citep{Hogan_2021}, where observations
are explicit Resource Description Framework (RDF) triples \citep{rdf12concepts,rdf12turtle} $(\texttt{head},\texttt{relation},\texttt{tail})$ and transfer
actions are directly inspectable. Prior work introduced
the RoomKG benchmark and established it as a test bed for memory-centric decision making under
partial observability \citep{kim2026temporalknowledgegraphmemorypartially}. Building on
that setting, we focus on \emph{short-term-to-long-term memory transfer}: deciding
keep/drop for each short-term observed triple before long-term insertion.

The key technical challenge is variable cardinality: the number of short-term items
changes across steps, so the transfer-action dimension is not fixed. We address this with a
shared-parameter per-item Q-learning design and a practical temporal-difference (TD) update over matched items
when consecutive short-term set sizes differ.

For controlled evaluation, we keep non-transfer components fixed. We compare against
symbolic baselines with temporal annotations, prior RoomKG history-based baselines (LSTM
\citep{10.1162/neco.1997.9.8.1735}/Transformer \citep{vaswani2023attentionneed}), and
two additional symbolic baselines (novel-only and random 50\% transfer).
At a long-term memory capacity of 128, learned transfer decisions outperform this suite. Within
the learned family, ablations over local/global and short-term-only/full-context settings show that a lightweight local short-term-only policy is strongest in this regime. We position this paper as a
focused mechanism-validation study of explicit memory transfer decisions, not a broad
benchmark sweep.

\section{Background and Problem Setup}
\label{sec:background}

\begin{figure}[t]
	\centering
	\begin{subfigure}[t]{0.49\linewidth}
		\centering
		\includegraphics[width=\linewidth]{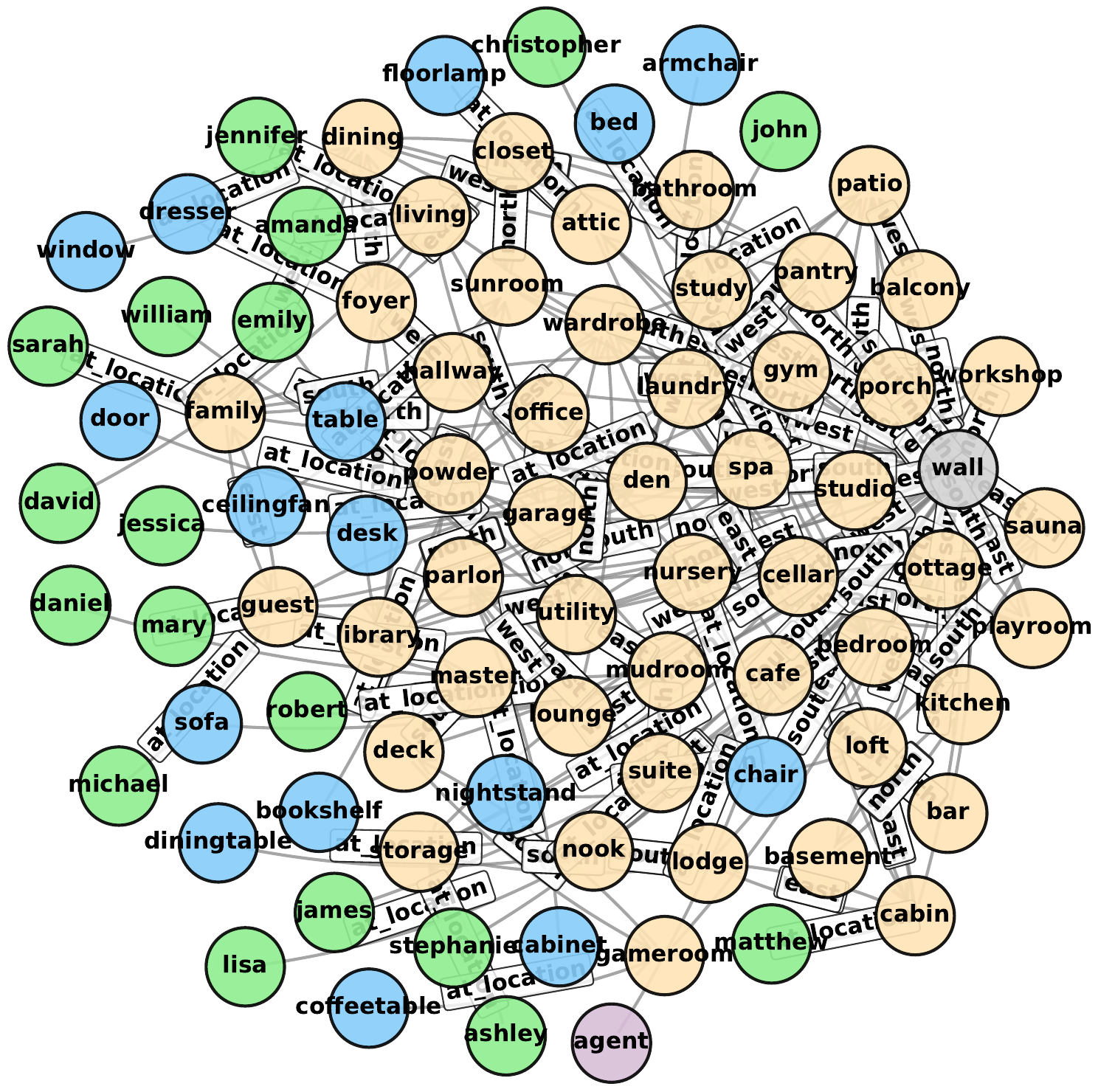}
		\caption{Hidden state ($s_{t=99}$), not directly observable to the agent.}
		\label{fig:hidden-state-step-099}
	\end{subfigure}
	\hfill
	\begin{subfigure}[t]{0.49\linewidth}
		\centering
		\includegraphics[width=\linewidth]{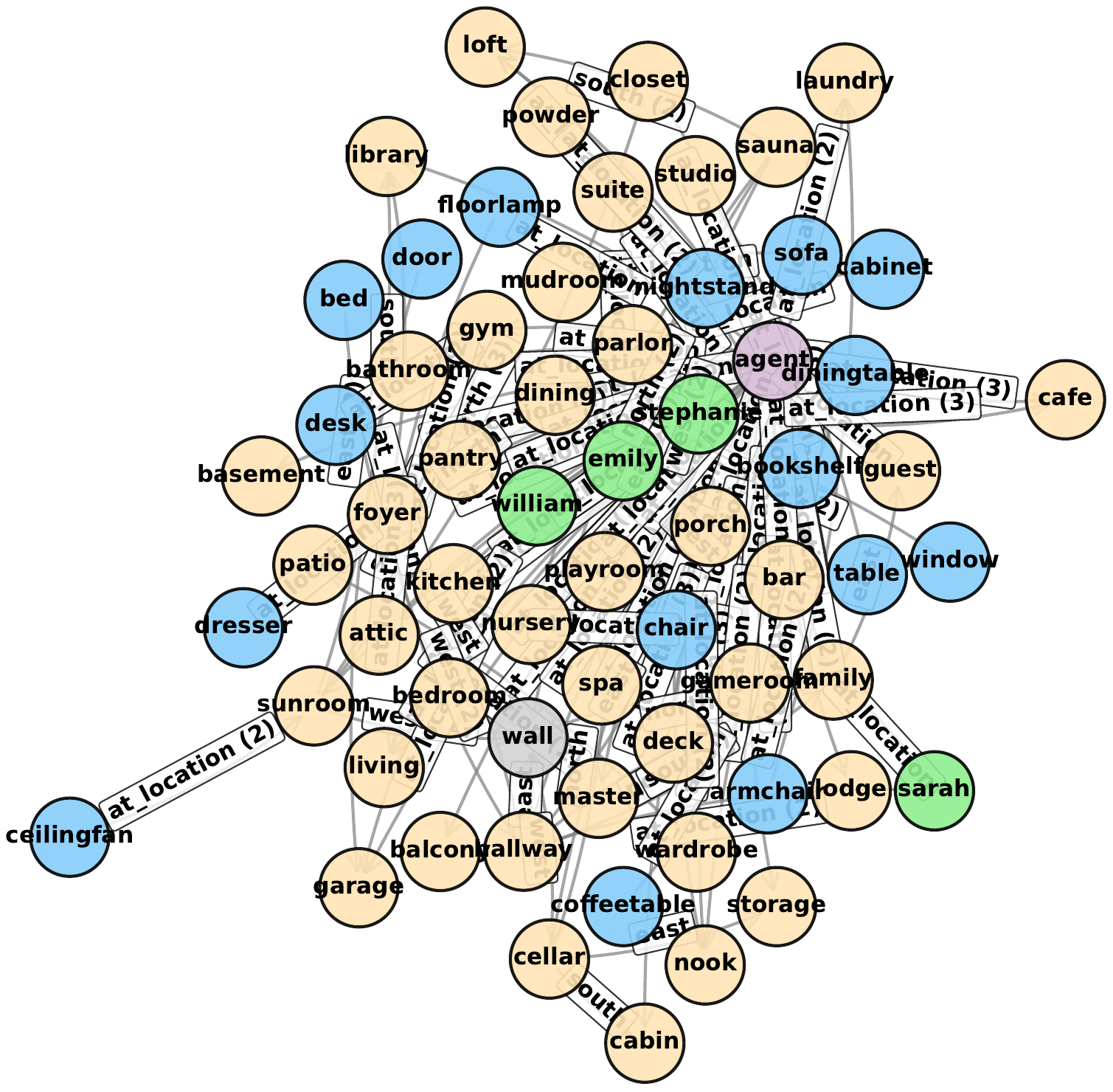}
		\caption{Our agent's internal symbolic memory state at step 99.}
		\label{fig:memory-state-step-099}
	\end{subfigure}
	\caption{Side-by-side comparison at step 99 in RoomKG: the environment hidden
	state (left) and our agent's internal symbolic memory state (right). Node colors indicate
	semantic categories: rooms (yellow), agent (purple), static objects (blue), moving
	objects (green), and walls (grey). In implementation, each memory is stored as a
	base triple together with temporal annotations. For
	readability in the memory-state plot, memories with the same main triple
	$(\texttt{head},\texttt{relation},\texttt{tail})$ are collapsed, and relation labels
	marked with $(N)$ indicate that $N$ distinct memories share that main triple
	(differing by temporal annotations).}
	\label{fig:state-memory-step-099}
\end{figure}

\subsection{Environment and Partial Observability Setup}
We consider the RoomKG benchmark, a partially observable sequential decision-making environment
where the full world state is hidden and the agent receives only local observations at
each step \citep{kim2026temporalknowledgegraphmemorypartially}. Let $s_t$ denote the
hidden state and $o_t$ the observation at time $t$. The environment is intentionally
designed around explicit graph structure: hidden states and observations are
expressed as RDF graphs
\citep{rdf12concepts,rdf12turtle}, and memory augments base
triples with temporal annotations, keeping contents explicit and auditable
while still supporting RL training.

Concretely, the hidden world graph includes rooms, objects, walls, and the agent, while
each $o_t$ is a local induced subgraph around the current room. Since single-step views
are incomplete, correct behavior requires integrating information across time through
memory rather than reacting to $o_t$ alone.

Unless stated otherwise, we use the standard RoomKG configuration from prior work:
a $7\times7$ grid ($49$ rooms), $18$ static objects, $18$ moving objects, $36$ inner
walls with periodic schedules, and $100$ steps per episode.
Following prior work, we use the standard \texttt{train}/\texttt{test} split:
\texttt{train} for policy learning/model selection and \texttt{test} as a held-out
evaluation environment with the same dynamics but different query ordering.

Figures~\ref{fig:hidden-state-step-099} and~\ref{fig:memory-state-step-099} show,
side by side, an example hidden-state snapshot ($s_t$) and the corresponding internal
symbolic memory-state view of our agent at the same step. This gap between hidden state and local
observation motivates explicit memory transfer decisions under partial observability.

\subsection{Memory Representation and Transfer-Decision Problem}
We model interaction as a partially observable control problem over
$\langle \mathcal{S},\mathcal{O},\mathcal{A},P,R,\Omega,\gamma\rangle$, where the agent
optimizes the discounted return.
Here, $\mathcal{S}$ is the hidden-state space, $\mathcal{O}$ the observation space,
$\mathcal{A}$ the action space, $P(s_{t+1}\mid s_t,a_t)$ the transition model,
$R(s_t,a_t)$ the reward function, $\Omega(o_t\mid s_t)$ the observation model, and
$\gamma\in[0,1)$ the discount factor.
The objective is
$J(\pi)=\mathbb{E}_{\pi}\!\left[\sum_{t=0}^{T-1} \gamma^t r_t\right]$, where $T$ is
the episode horizon and $r_t$ is the reward at step $t$.
The key design question is how to decide what to transfer to long-term memory under
partial observability and finite capacity.

We use the standard POMDP tuple $\langle \mathcal{S},\mathcal{O},\mathcal{A},P,R,\Omega,\gamma\rangle$
to describe the hidden environment dynamics. The transfer controller
operates on top of that environment as a structured internal decision process induced by
the current short-term memory. If the current short-term set contains
$n_t:=|\mathcal{M}^{\mathrm{short}}_t|$ items, then the transfer decision space at step
$t$ is
$\mathcal{A}^{\mathrm{tr}}_t:=\{0,1\}^{n_t}$, the set of binary keep/drop assignments with
one entry per short-term item. This makes explicit that the variable-cardinality
keep/drop decisions are not a fixed discrete action set of the underlying environment,
but an internal fact-selection mechanism conditioned on the current observation set.

In RoomKG, reward is emitted only through question-answering correctness, so memory
decisions are learned through their downstream effect on future QA outcomes. Intuitively,
keep/drop transfer decisions shape the agent's effective internal state: not the
true hidden state $s_t$, but a belief-like symbolic memory state built from partial
observations. Better transfer decisions should produce memory states that more closely track
task-relevant aspects of $s_t$, which in turn improves decision quality and return.

At each step $t$, the agent receives a short-term set
$\mathcal{M}^{\mathrm{short}}_t=\{m_{t,1},\dots,m_{t,n_t}\}$, where $n_t$ varies over time,
and maintains a capacity-limited long-term memory $\mathcal{M}^{\mathrm{long}}_t$ with
$|\mathcal{M}^{\mathrm{long}}_t|\le K$. The keep/drop policy outputs one binary action
per short-term item, $a^{\mathrm{tr}}_{t,i}\in\{0,1\}$, deciding whether $m_{t,i}$ is
kept (transferred/refreshed in long-term memory) or dropped.

We treat the agent's effective internal memory state as
$M_t = \mathcal{M}^{\mathrm{short}}_t \cup \mathcal{M}^{\mathrm{long}}_t$, where
$\mathcal{M}^{\mathrm{short}}_t$ captures immediate local observations and
$\mathcal{M}^{\mathrm{long}}_t$ captures persistent, capacity-limited memory. This
internal symbolic state is not the hidden state $s_t$; rather, it is a belief-like
task state built from partial observations and memory updates.

Its evolution is induced by the next observation, transfer decisions, and fixed
non-transfer policies. Writing the per-step transfer assignment as
$\mathbf{a}^{\mathrm{tr}}_t=(a^{\mathrm{tr}}_{t,1},\dots,a^{\mathrm{tr}}_{t,n_t})$, we can
view the update abstractly as
\[
M_{t+1}=U\!\left(M_t,o_{t+1},\mathbf{a}^{\mathrm{tr}}_t\right),
\]
where $U$ denotes deterministic short-term refresh, selective insertion into
long-term memory, and any resulting capacity handling under the fixed eviction rule.

This yields a variable-cardinality action structure: the number of transfer decisions at
time $t$ equals $n_t$. We therefore study memory transfer as a focused RL problem over
symbolic memory items, while holding non-transfer components fixed so that observed
performance differences can be attributed to transfer decisions themselves.

\section{Method}
\label{sec:method}

\subsection{Method Overview and Scope}
Our method learns only the transfer keep/drop policy. At each step, the agent receives
the current symbolic memory state $M_t$ and outputs one binary decision per short-term
item in $\mathcal{M}^{\mathrm{short}}_t$: keep (transfer to long-term memory) or drop. To
isolate the effect of transfer decisions, non-transfer components (question answering,
exploration, and eviction) are held fixed by symbolic heuristics during this method
stage.

The learning signal remains task-level reward from the environment, so the keep/drop
policy is optimized through delayed credit assignment: transfer decisions are judged by
their downstream contribution to future QA correctness.

A naive alternative would treat transfer decisions as one joint combinatorial action over
all short-term items. If there are $n_t$ items at step $t$, this yields $2^{n_t}$
actions, which becomes quickly impractical and also assumes a fixed action indexing
across time. Here, short-term memory is a set (not a sequence), so its cardinality and
ordering are not stable decision coordinates. This motivates a set-compatible,
per-item factorization rather than a single monolithic action head.

\subsection{Per-Item Q-Value Parameterization}
Using $n_t:=|\mathcal{M}^{\mathrm{short}}_t|$, we use a shared-parameter function
$Q_\theta$ that maps the current symbolic state to per-item action values:
\[
Q_\theta(M_t)\;\rightarrow\;\{\mathbf{q}_{t,i}\}_{i=1}^{n_t},
\quad
\mathbf{q}_{t,i}\in\mathbb{R}^{2}.
\]
Each row $\mathbf{q}_{t,i}$ contains Q-values for the two actions $a\in\{0,1\}$ (keep
vs. drop for item $i$). Shared parameters across items allow one model to score
variable-length short-term sets without requiring a fixed action count. Here $i$ is a
step-local index over the current short-term set. Because $\mathcal{M}^{\mathrm{short}}_t$
is a set, it has no canonical order; the environment therefore emits its items in a
random order at each step. Thus, $i$ does not identify the same memory item across time
steps. Equivalently, when we write $Q_\theta(M_t,i,a)$ below, we mean the entry of the
$i$-th output row indexed by action $a$, that is,
$Q_\theta(M_t,i,a)=[\mathbf{q}_{t,i}]_a$.

In implementation, $M_t$ is converted to a graph
$G_t=(V_t,E_t,R_t,\mathcal{Q}_t)$, where $\mathcal{Q}_t$ denotes temporal annotation features,
and passed through a graph neural network (GNN) encoder.
Let $\mathbf{h}^{(0)}_v$ and $\mathbf{r}^{(0)}_p$ be initial entity and relation
embeddings. After $L$ message-passing layers:
\[
\big\{\mathbf{h}^{(L)}_v\big\}_{v\in V_t},\;\big\{\mathbf{r}^{(L)}_p\big\}_{p\in R_t}
=\mathrm{GNN}_\phi\!\left(G_t,\{\mathbf{h}^{(0)}_v\},\{\mathbf{r}^{(0)}_p\}\right).
\]
For each short-term memory item
$m_{t,i}=(u_{t,i},p_{t,i},v_{t,i},\xi_{t,i})\in\mathcal{M}^{\mathrm{short}}_t$,
we build the item representation by concatenating the updated endpoint embeddings:
\[
\mathbf{z}_{t,i}=\big[\mathbf{h}^{(L)}_{u_{t,i}}\,\|\,\mathbf{h}^{(L)}_{v_{t,i}}\big].
\]
Because these node embeddings are produced by message passing on $G_t$,
$\mathbf{z}_{t,i}$ already contains contextual information from neighboring short-term
and long-term memory triples included in the encoder input. Thus, the final per-item
transfer representation explicitly concatenates only head and tail node embeddings;
relation and temporal-annotation effects are incorporated implicitly through the GNN
updates that produced those node embeddings. Q-values are then produced by a small MLP
head:
\[
\mathbf{q}_{t,i}=\mathrm{MLP}_{\mathrm{tr}}(\mathbf{z}_{t,i})\in\mathbb{R}^{2}.
\]

This design is related in spirit to independent Q-learning
\citep{10.5555/3091529.3091572}, because action values are
computed per decision unit. However, unlike independent multi-agent formulations, we do
not optimize separate MDPs: all per-item decisions are components of one agent solving
the same underlying task MDP, and all components share parameters $\theta$.

Action selection is per-item $\epsilon$-greedy. For each $i$,
$a_{t,i}=\arg\max_{a\in\{0,1\}} Q_\theta(M_t,i,a)$ with probability $1-\epsilon$, and a
uniform random action otherwise.

\subsection{TD Learning with Variable-Cardinality Matching}
Transitions are stored in replay as
$(M_t,\mathbf{a}_t,r_t,M_{t+1},d_t)$, where $\mathbf{a}_t$ is the vector of per-item
transfer actions and $d_t\in\{0,1\}$ is terminal status. Since
$|\mathcal{M}^{\mathrm{short}}_t|$ and $|\mathcal{M}^{\mathrm{short}}_{t+1}|$ can differ, we
compute TD updates on matched items only.

Because $\mathcal{M}^{\mathrm{short}}_t$ is a set, there is no canonical item order for
cross-step alignment. We therefore use a randomized, order-agnostic matching strategy in
practice: observations are shuffled at environment emission time, replay samples
transitions stochastically, and TD updates pair items index-wise up to the shared length
$\ell_b$. This avoids imposing an artificial sequence semantics on inherently set-valued
memory.

For sampled transition $b$, let
$\ell_b=\min\!\big(n_b,n'_b\big)$ with
$n_b=|\mathcal{M}^{\mathrm{short}}_b|$ and
$n'_b=|\mathcal{M}^{\mathrm{short}}_{b+1}|$.
For each matched index $j\in\{1,\dots,\ell_b\}$, where $j$ denotes only the $j$-th
sampled pair under this stochastic procedure and not the same persistent fact across
consecutive steps:
\[
y_{b,j}=r_b+\gamma(1-d_b)\,\hat{q}_{b+1,j},
\]
where $\hat{q}_{b+1,j}=\max_{a\in\{0,1\}} Q_{\bar{\theta}}(M_{b+1},j,a)$. The current estimate is
$q_{b,j}=Q_\theta(M_b,j,a_{b,j})$.

We minimize the expected squared TD error over replay samples and matched indices:
\[
\mathcal{L}(\theta)=\mathbb{E}_{(M_t,\mathbf{a}_t,r_t,M_{t+1},d_t)\sim\mathcal{D},\,\text{match}}
\left[\frac{1}{\ell_t}\sum_{j=1}^{\ell_t}\left(q_{t,j}-y_{t,j}\right)^2\right],
\]
with $\ell_t=\min\!\big(|\mathcal{M}^{\mathrm{short}}_t|,|\mathcal{M}^{\mathrm{short}}_{t+1}|\big)$.
This keeps learning stable when short-term cardinality changes across steps while
preserving dense supervision from each decision point. The expectation is thus over both
sampled replay transitions and the stochastic emission/matching order. Normalizing by
$\ell_t$ keeps the per-transition loss on a comparable scale when different transitions
yield different numbers of matched items.

Each training iteration follows: (1) select per-item transfer actions by
$\epsilon$-greedy policy, (2) execute environment step and obtain reward, (3) store
transition in replay, (4) sample mini-batch after warm start, (5) update online network
by TD loss, and (6) periodically hard-update the target network. This is standard
off-policy DQN training, specialized to variable-cardinality symbolic transfer decisions
\citep{mnih2013playingatarideepreinforcement}.

\section{Experimental Setup}
\label{sec:experimental-setup}

\subsection{Experimental Protocol}
We evaluate on the RoomKG benchmark with the standard \texttt{train}/\texttt{test}
split: \texttt{train} is used for learning/model selection and \texttt{test} is a
held-out environment with the same dynamics but different query ordering
\citep{kim2026temporalknowledgegraphmemorypartially}. Unless stated otherwise, we use
the default configuration from prior work (grid length $7$, $49$ rooms, $18$ static
objects, $18$ moving objects, $36$ periodic inner walls, and $100$ steps per episode)
and focus on a long-term memory capacity of 128. We choose the RoomKG benchmark and this
long-term memory capacity as a focused setup where symbolic keep/drop decisions are
directly auditable and memory pressure remains strong enough for short-term-to-long-term
transfer to affect downstream QA performance. We view this as a controlled mechanism
study rather than a broad robustness claim across environments or capacities.

\subsection{Compared Methods and Metrics}
We compare learned transfer decisions against a baseline suite consisting of:
(i) baselines from prior RoomKG work, including simple symbolic systems with temporal annotations
and end-to-end history-based neural baselines (LSTM
\citep{10.1162/neco.1997.9.8.1735} and Transformer \citep{vaswani2023attentionneed}), and
(ii) additional symbolic transfer baselines introduced here (Novel-Only and
Random-Transfer ($p=0.5$)). For the learned transfer family, we additionally run an ablation
grid over GNN encoder choices and transfer-policy modes. Encoder choices include
graph convolutional networks (GCN) \citep{kipf2017semisupervisedclassificationgraphconvolutional},
relational graph convolutional networks (R-GCN) to inject relation-type-specific message passing
\citep{schlichtkrull2017modelingrelationaldatagraph}, and
StarE hyper-relational graph neural networks (StarE-GNN) to incorporate annotation features
\citep{galkin2020messagepassinghyperrelationalknowledge}.
Transfer-policy modes are:
\emph{Local-Full} (per-item learned transfer decisions with full memory context),
\emph{Local-STM} (per-item learned transfer decisions with short-term-only encoder input),
\emph{Global-Full} (one pooled learned decision vector shared across short-term items using full context),
and \emph{Global-STM} (pooled shared learned decisions with short-term-only encoder input).

In all comparisons, non-transfer components are kept fixed so performance differences
reflect transfer decisions. For the symbolic temporal-annotation baselines, these fixed
components use most-recently-used (MRU) question answering, MRU exploration, and
least-recently-used (LRU) eviction, computed from temporal annotations attached to
memory entries (e.g., \texttt{time\_added}, \texttt{last\_accessed},
\texttt{num\_recalled}); see Appendix~\ref{sec:appendix-policy-definitions} for the
precise symbolic policy definitions and annotation conventions.

The primary metric is episode-level question-answering score (number of correctly
answered queries per episode). We report mean and standard deviation across random seeds
for both \texttt{train} and \texttt{test}, and use the held-out \texttt{test} split as
the main generalization indicator. We use $n{=}5$ seeds as a transparent small-scale
evaluation budget and report mean $\pm$ standard deviation. Baselines follow the RoomKG
setup and aligned default training settings used in prior work where applicable; within
our compute budget we kept training budgets broadly comparable across learned variants
rather than performing architecture-specific exhaustive hyperparameter sweeps for every
baseline. The implementation used in this paper is open sourced at
\url{https://github.com/humemai/kg-memory-transfer}.

\section{Results}
\label{sec:results}

\subsection{Quantitative Results}
Table~\ref{tab:main-results-k128} reports both \texttt{train} and \texttt{test}
performance at a long-term memory capacity of 128 (mean $\pm$ std over $n{=}5$ seeds for
each setting). The best result is obtained by the learned transfer-policy configuration
with GCN encoder and \emph{Local-STM} transfer policy.

Learned transfer decisions outperform end-to-end LSTM/Transformer baselines in both
\texttt{train} and \texttt{test}, and also improve over symbolic baselines with temporal
annotations on held-out performance with stable variance. Within learned variants, local
per-item transfer decisions outperform pooled global decisions, and short-term-only
input outperforms full-context input in this regime.

Quantitatively, the best setting (GCN + \emph{Local-STM}) reaches $38.920\pm3.090$ on
\texttt{test}, compared with $31.960\pm1.255$ for the strongest symbolic baseline with
temporal annotations (\emph{Novel-Only}), a gain of $+6.96$ points. Against end-to-end
neural baselines, the margin is larger: $+27.12$ over Transformer and $+31.32$ over LSTM
on \texttt{test}. This places the main improvement in explicit transfer decisions rather
than in replacing symbolic memory with sequence-only models.

Within learned variants, architecture and transfer mode have very different effects.
Keeping the same GCN encoder, \emph{Local-STM} outperforms \emph{Local-Full} ($38.920$
vs. $35.080$) and both global variants ($19.440$ and $9.080$), indicating that per-item
local transfer is more effective than pooled global decisions in this regime. Across
encoder families under \emph{Local-STM}, GCN is markedly stronger than R-GCN and
StarE-GNN in this setup, suggesting that extra relational/annotation modeling does not
automatically translate into higher QA reward for the present memory-transfer task.

\begin{table}[t]
\centering
\small
\caption{Main \texttt{train}/\texttt{test} results at long-term memory capacity 128.
Scores are episode-level QA accuracy (higher is better), reported as mean $\pm$ std
over 5 seeds. Variant names: \emph{Always-Transfer} transfers every short-term item,
\emph{Novel-Only} transfers only items not already in long-term memory,
\emph{Random-Transfer (p=0.5)} transfers each short-term item with probability $0.5$,
\emph{Local-Full} and \emph{Local-STM} are per-item learned transfer policies,
and \emph{Global-Full} and \emph{Global-STM} are pooled shared-decision learned transfer policies.
End-to-end DQN rows use \emph{Always-Transfer} memory transfer.}
\label{tab:main-results-k128}
\begin{tabular}{l l c c}
\hline
Family & Variant & Train score & Test score \\
\hline
Symbolic temporal-RDF & Novel-Only & $32.120 \pm 1.204$ & $31.960 \pm 1.255$ \\
Symbolic temporal-RDF & Always-Transfer & $28.600 \pm 1.702$ & $29.680 \pm 2.317$ \\
Symbolic temporal-RDF & Random-Transfer ($p=0.5$) & $21.480 \pm 2.282$ & $22.440 \pm 3.567$ \\
End-to-end DQN & Transformer + Always-Transfer & $16.800 \pm 0.748$ & $11.800 \pm 3.187$ \\
End-to-end DQN & LSTM + Always-Transfer & $10.200 \pm 1.166$ & $7.600 \pm 2.332$ \\
\hline
DQN temporal-RDF & GCN + Local-STM & $\mathbf{37.240 \pm 1.713}$ & $\mathbf{38.920 \pm 3.090}$ \\
DQN temporal-RDF & R-GCN + Local-STM & $12.920 \pm 4.657$ & $13.600 \pm 3.132$ \\
DQN temporal-RDF & StarE-GNN + Local-STM & $13.680 \pm 2.847$ & $13.760 \pm 3.032$ \\
DQN temporal-RDF & GCN + Local-Full & $34.800 \pm 1.523$ & $35.080 \pm 3.037$ \\
DQN temporal-RDF & GCN + Global-Full & $18.240 \pm 8.306$ & $19.440 \pm 10.506$ \\
DQN temporal-RDF & GCN + Global-STM & $8.880 \pm 3.000$ & $9.080 \pm 3.645$ \\
\hline
\end{tabular}
\end{table}

\subsection{What the Learned Transfer Policy Does}
\label{sec:results-transfer-behavior}

Unless stated otherwise, the analyses in Sections~\ref{sec:results-transfer-behavior}
and~\ref{sec:results-qualitative-steps} are taken from a single trained \textbf{DQN
temporal-RDF (GCN + Local-STM)} model (the best configuration in
Table~\ref{tab:main-results-k128}) on held-out \texttt{test} episodes.

\begin{figure}[t]
\centering
\includegraphics[width=0.75\linewidth]{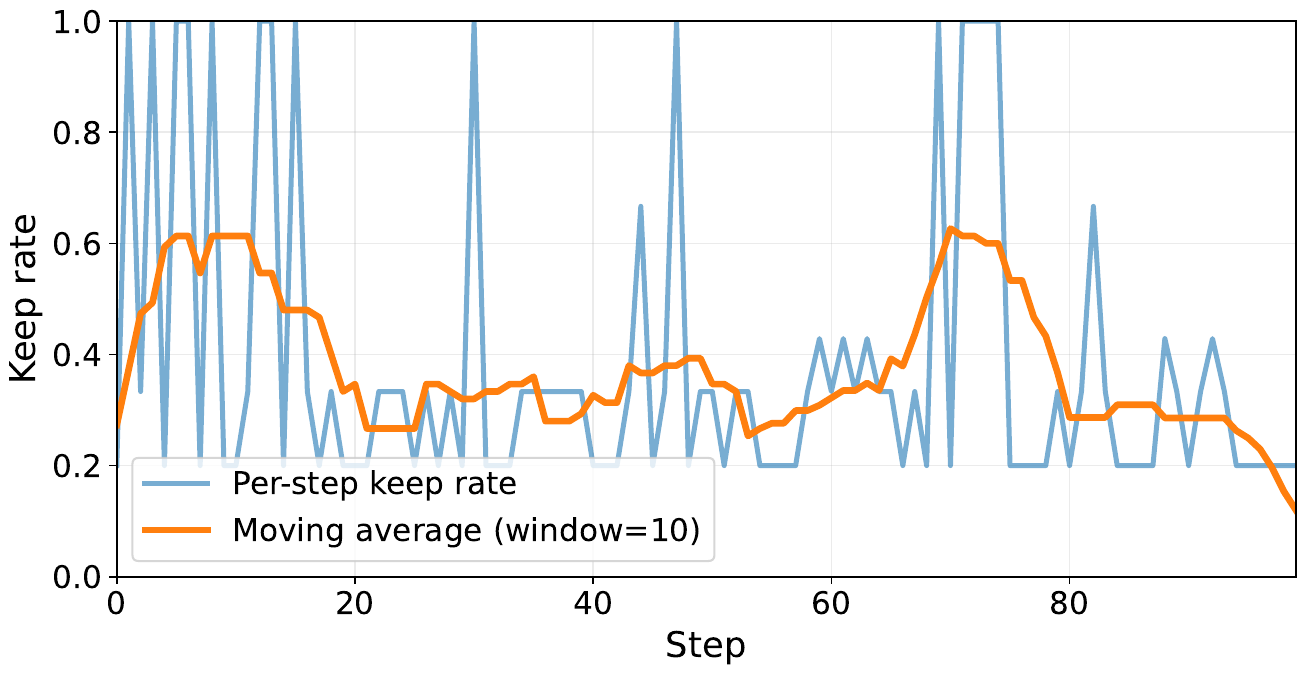}
\caption{Keep-rate trajectory over held-out \texttt{test} steps (higher means
more short-term items are transferred to long-term memory). The moving average highlights
that transfer is adaptive rather than fixed-rate.}
\label{fig:keep-rate-over-time}
\end{figure}

To understand what the learned transfer policy actually retains, we analyze per-step
keep-or-drop transfer decisions on held-out \texttt{test} episodes across 560 transfer
decisions (100 steps; about 5--6 candidate short-term memories per step). The policy
keeps 224 items and drops 336 items (overall keep rate $0.40$).
Figure~\ref{fig:keep-rate-over-time} shows the global trajectory of this keep/drop
behavior over time.

Three stable patterns emerge. First, the policy strongly preserves the agent's own
position signal: \texttt{(agent, at\_location, room)} is kept 98/100 times. This is not
just a preference; in this environment it appears functionally necessary. Without
reliable retention of the agent's current location and visited-room trace, the agent
policy cannot localize itself or run the internal BFS traversal over the memory-built
graph map toward unexplored rooms. Second, the policy retains question-relevant object
locations: for objects appearing in the held-out query set, \texttt{at\_location} facts
are kept 58 times and dropped only 2 times. Third, the policy usually drops
room-direction map links (for example,
\texttt{north}/\texttt{south}/\texttt{east}/\texttt{west} links): it keeps 68 and drops
332. This behavior is similar for links to walls and links to other rooms, suggesting
that many map-link facts are treated as noisy or less important under changing layouts.

Figure~\ref{fig:memory-state-step-099} complements this action-level view by showing the
resulting internal memory snapshot at the final step. Together, the decision logs and
the step-99 memory state indicate a selective memory policy: keep location facts that
directly support navigation and question answering, while dropping many room-direction
map links (for example, \texttt{north}/\texttt{south}/\texttt{east}/\texttt{west}
edges).

\subsection{Qualitative Step-Level Examples}
\label{sec:results-qualitative-steps}

At step 8, the queried object is \texttt{john}, and the policy keeps \texttt{(john,
at\_location, playroom)} (together with \texttt{(agent, at\_location, playroom)}). In
this case it also retains the full locally observed playroom neighborhood, showing that
the policy does not simply keep isolated object-location facts: at some steps it
preserves the co-observed room context alongside the queried object and the agent's
location. At step 26, the policy keeps \texttt{(agent, at\_location, studio)} and
\texttt{(table, at\_location, studio)} while dropping four directional links from
\texttt{studio}; this more directly illustrates the selective pattern highlighted by our
aggregate analysis, namely preserving self-localization and object-location facts while
compressing many map-direction links. At step 62, the queried object is
\texttt{william}, and the policy keeps \texttt{(william, at\_location, living)} while
dropping four directional links from \texttt{living}; this again matches the pattern of
preserving likely QA anchors while compressing many map-direction links.

An additional benefit is interpretability: because memory transfers are symbolic
(explicit RDF triples with temporal annotations), we can directly inspect what is kept
or dropped at each step. This is much harder in latent-memory approaches where internal
state updates are distributed in hidden vectors
\citep{hausknecht2017deeprecurrentqlearningpartially,graves2014neuralturingmachines,Graves2016HybridCU}.

\section{Related Work}
\label{sec:related-work}

\textbf{Memory in RL under partial observability.}
Classic approaches to partial observability frequently rely on latent memory, including
recurrent value-based agents and neural external-memory architectures
\citep{hausknecht2017deeprecurrentqlearningpartially,esslinger2022deeptransformerqnetworks,graves2014neuralturingmachines,Graves2016HybridCU,pritzel2017neuralepisodiccontrol}.
These methods are strong at sequence modeling, but memory updates are hard to inspect at
fact level. More broadly, batch RL under partial observability also raises stability
concerns such as overfitting and asymptotic bias
\citep{francoislavet2019overfittingasymptoticbiasbatch}. We target this gap with
explicit keep/drop transfer decisions for each short-term symbolic fact.

Recent POMDP studies have also explored compact history summarization and explicit
memory decisions, including memory traces as an alternative to fixed windows
\citep{eberhard2025memorytraces} and external-memory action formulations where the agent
jointly chooses environment and memory-transfer actions
\citep{icarte2020actremembering}. Our formulation is closest in spirit to this line, but
specializes it to symbolic triple-level updates in temporal KG memory with step-level
inspectability.

\textbf{Symbolic and knowledge-graph memory for RL.}
The RoomKG benchmark introduced a temporal knowledge-graph memory setting that exposes
the memory-management challenge under partial observability
\citep{kim2026temporalknowledgegraphmemorypartially}. More broadly, RDF and
knowledge-graph formalisms provide structured, auditable representations
\citep{Hogan_2021}. In our setting, temporal triple annotations keep stored content
explicit and inspectable without changing the RL focus of the problem. This symbolic
view differs from latent sequence memory in that stored content remains semantically
typed and queryable. Earlier work by \citet{kim2023machineshorttermepisodicsemantic}
modeled a broader agent with short-term, episodic, and semantic knowledge-graph memory
systems; our paper isolates one specific mechanism in that design space, namely
short-term-to-long-term transfer under fixed downstream policies. The central policy
question is which observed facts to transfer under capacity constraints.

Related neuro-symbolic RL work has combined neural function approximation with symbolic
structure or rules, including symbolic priors for RL, symbolic policy discovery, and
hierarchical neuro-symbolic designs
\citep{garcez2018symbolicreinforcementlearningcommon,pmlr-v139-landajuela21a,ijcai2022p742}.
Recent studies also study neuro-symbolic guidance and action-constraint mechanisms for
improving sample efficiency in deep RL
\citep{veronese2026sampleefficientneurosymbolic,han2026neurosymbolicactionmasking}. Our
setting is narrower: explicit triple-level transfer decisions for temporal KG memory
under partial observability, with step-level inspectability.

\textbf{Graph encoders for relational state representations.}
Our experiments use standard graph-neural backbones for relational encoding---GCN,
R-GCN, and StarE-GNN
\citep{kipf2017semisupervisedclassificationgraphconvolutional,schlichtkrull2017modelingrelationaldatagraph,galkin2020messagepassinghyperrelationalknowledge}.
They provide a controlled test bed for transfer-policy learning; our claim is about
memory transfer decisions, not a new GNN.

\textbf{Transfer decisions and model capacity.}
Most prior neural baselines in this setting either transfer by fixed rules or rely on
end-to-end sequence memory. In contrast, we formulate short-term-to-long-term transfer
as an RL decision problem over variable-cardinality short-term sets, trained with a
practical off-policy temporal-difference (TD) recipe inspired by DQN-style learning
\citep{mnih2013playingatarideepreinforcement}. The key technical focus is
variable-cardinality transfer decisions: the number of short-term items changes across
steps, so transfer decisions must be produced and trained robustly without a fixed
action dimension. Compared with fixed symbolic transfer heuristics, our approach learns
transfer decisions; compared with latent-memory baselines, it retains fact-level
interpretability while improving transfer quality under memory limits.

\section{Conclusion}
\label{sec:conclusion}

We studied short-term-to-long-term memory transfer as a first-class RL decision in a
partially observable temporal knowledge-graph setting. Instead of relying on opaque
latent-state updates or fixed symbolic transfer rules, our method learns explicit
keep/drop actions for each observed triple under memory constraints. Empirically, the
learned policy improves QA over symbolic and neural baselines at long-term memory
capacity 128, and the action-level analysis explains why: it preserves navigation- and
query-critical facts while discarding many lower-value candidate facts. This supports
our central claim that interpretable memory transfer decisions are effective in
memory-constrained POMDP RL, while also highlighting scope limits: evaluation is limited
to RoomKG and one main memory-capacity regime, non-transfer components are fixed for
attribution, and temporal-difference (TD) matching for variable-cardinality short-term
sets is practical rather than theoretically complete. Future work should test broader
environments/capacities, stronger memory baselines, and joint end-to-end training that
preserves fact-level interpretability.

\section*{Acknowledgments}
\label{sec:ack}
This research was (partially) funded by the Hybrid Intelligence Center, a 10-year
program funded by the Dutch Ministry of Education, Culture and Science through the
Netherlands Organization for Scientific Research,
\url{https://www.hybrid-intelligence-centre.nl/}.


\bibliography{main}
\bibliographystyle{rlj}

\beginSupplementaryMaterials

\section{Policy Definitions Used in Experiments}
\label{sec:appendix-policy-definitions}

This section defines the symbolic non-transfer policies used while learning and evaluating
transfer decisions.

\paragraph{Question-answering policy.}
Given an RDF/SPARQL-style query triple \citep{rdf12concepts,sparql12query} with one
missing element (e.g., \texttt{(john, at\_location, ?)} $\rightarrow$ missing tail), the
agent ranks matching candidates in memory using temporal annotations such as recency and
frequency and returns the top-ranked answer. \texttt{mra} denotes
\emph{most-recently-added} (rank by \texttt{time\_added}), \texttt{mru} denotes
\emph{most-recently-used} (rank by \texttt{last\_accessed}), and \texttt{mfu} denotes
\emph{most-frequently-used} (rank by \texttt{num\_recalled}).

\paragraph{Exploration policy.}
The agent uses memory-filtered graph search for navigation. Annotation-based ranking
resolves conflicting room/edge memories, then BFS is run on the resulting graph map; the
environment action is the first move on the selected path.

\paragraph{Eviction policy.}
When long-term memory reaches capacity, one memory entry is evicted.
\texttt{fifo} denotes \emph{first-in-first-out} (evict oldest inserted entry),
\texttt{lru} denotes \emph{least-recently-used} (evict minimum
\texttt{last\_accessed}), and \texttt{lfu} denotes \emph{least-frequently-used}
(evict minimum \texttt{num\_recalled}).

\paragraph{Transfer policy.}
Main baselines are \emph{Always-Transfer} (transfer every short-term item),
\emph{Novel-Only} (transfer only items not already present in long-term memory), and
\emph{Random-Transfer (p=0.5)}. Learned variants are \emph{Local-Full},
\emph{Local-STM}, \emph{Global-Full}, and \emph{Global-STM}.

\section{Formal Training Algorithm (Transfer Decisions)}

\noindent\textbf{Algorithm 1: Per-item DQN training with variable-cardinality matching}

\begin{enumerate}
	\item Initialize online parameters $\theta$, target parameters $\bar\theta\leftarrow\theta$, replay buffer $\mathcal{D}$.
	\item For each environment step $t$:
	\begin{enumerate}
		\item Build current memory state $M_t=(\mathcal{M}^{\mathrm{short}}_t,\mathcal{M}^{\mathrm{long}}_t)$.
		\item For each short-term item $i\in\{1,\dots,n_t\}$, choose
		$a_{t,i}$ by $\epsilon$-greedy from $Q_\theta(M_t,i,\cdot)$.
		\item Execute transfer decisions, receive reward $r_t$, next state $M_{t+1}$, and terminal flag $d_t$.
		\item Store transition $(M_t,\mathbf{a}_t,r_t,M_{t+1},d_t)$ in $\mathcal{D}$.
		\item If replay warm start is reached, sample minibatch transitions from $\mathcal{D}$.
		\item For each sampled transition $b$, set
		$\ell_b=\min(|\mathcal{M}^{\mathrm{short}}_b|,|\mathcal{M}^{\mathrm{short}}_{b+1}|)$,
				then sample a stochastic step-local matching and pair matched items index-wise for
				$j=1,\dots,\ell_b$ and compute
		\[
			y_{b,j}=r_b+\gamma(1-d_b)\,\hat q_{b+1,j},
			\quad
			q_{b,j}=Q_\theta(M_b,j,a_{b,j}).
		\]
		where $\hat q_{b+1,j}=\max_{a\in\{0,1\}} Q_{\bar\theta}(M_{b+1},j,a)$.
		\item Minimize
		\[
			\mathcal{L}(\theta)=
			\mathbb{E}_{\text{replay},\,\text{match}}\left[\frac{1}{\ell_b}\sum_{j=1}^{\ell_b}(q_{b,j}-y_{b,j})^2\right].
		\]
		The expectation is over both replay sampling and the stochastic matching.
		\item Every fixed interval, update target network: $\bar\theta\leftarrow\theta$.
	\end{enumerate}
\end{enumerate}

\section{Hyperparameters and Protocol}

\begin{table}[tb]
\centering
\small
\caption{Core training settings used for the reported long-term memory capacity 128 experiments.}
\label{tab:appendix-hparams-core}
\begin{tabular}{l l}
\hline
Setting & Value \\
\hline
Episode horizon & 99 ($100$ steps/episode) \\
Training episodes & 200 \\
Training iterations & 20,000 \\
Replay buffer size & 20,000 \\
Warm start & 2,000 \\
Batch size & 32 \\
Target update interval & 50 \\
$\epsilon$ schedule & max $1.0$ to min $0.01$ over 10,000 iterations \\
Discount factor $\gamma$ & 0.95 \\
Learning rate & $10^{-4}$ \\
Double DQN & True \\
Gradient clipping & True (clip value 10.0) \\
Seeds & $\{0,5,10,15,20\}$ \\
\hline
\end{tabular}
\end{table}

\begin{table}[tb]
\centering
\small
\caption{Model-size settings used for the main long-term memory capacity 128 comparisons.}
\label{tab:appendix-hparams-arch}
\begin{tabular}{l l}
\hline
Component & Value \\
\hline
GCN / R-GCN / StarE embedding dim & 16 \\
GCN / R-GCN / StarE layers & 2 \\
R-GCN num bases & 20 \\
MLP hidden layers & 1 \\
\hline
\end{tabular}
\end{table}

For the best-performing GCN + Local-STM model in this paper, the total number of
trainable parameters is \textbf{8,339}, indicating a relatively compact
model.

\section{Runtime and Compute}

All experiments were run on CPU only: \textbf{AMD Ryzen 9 7950X (16-Core Processor)}
with \textbf{128GB DDR5 RAM}.
The best-performing \textbf{DQN temporal-RDF GCN + Local-STM} model required about
\textbf{9 hours} of training on this hardware.

\section{Additional Hidden-State and Memory Snapshots}

\begin{figure}[tb]
\centering
\includegraphics[width=\linewidth]{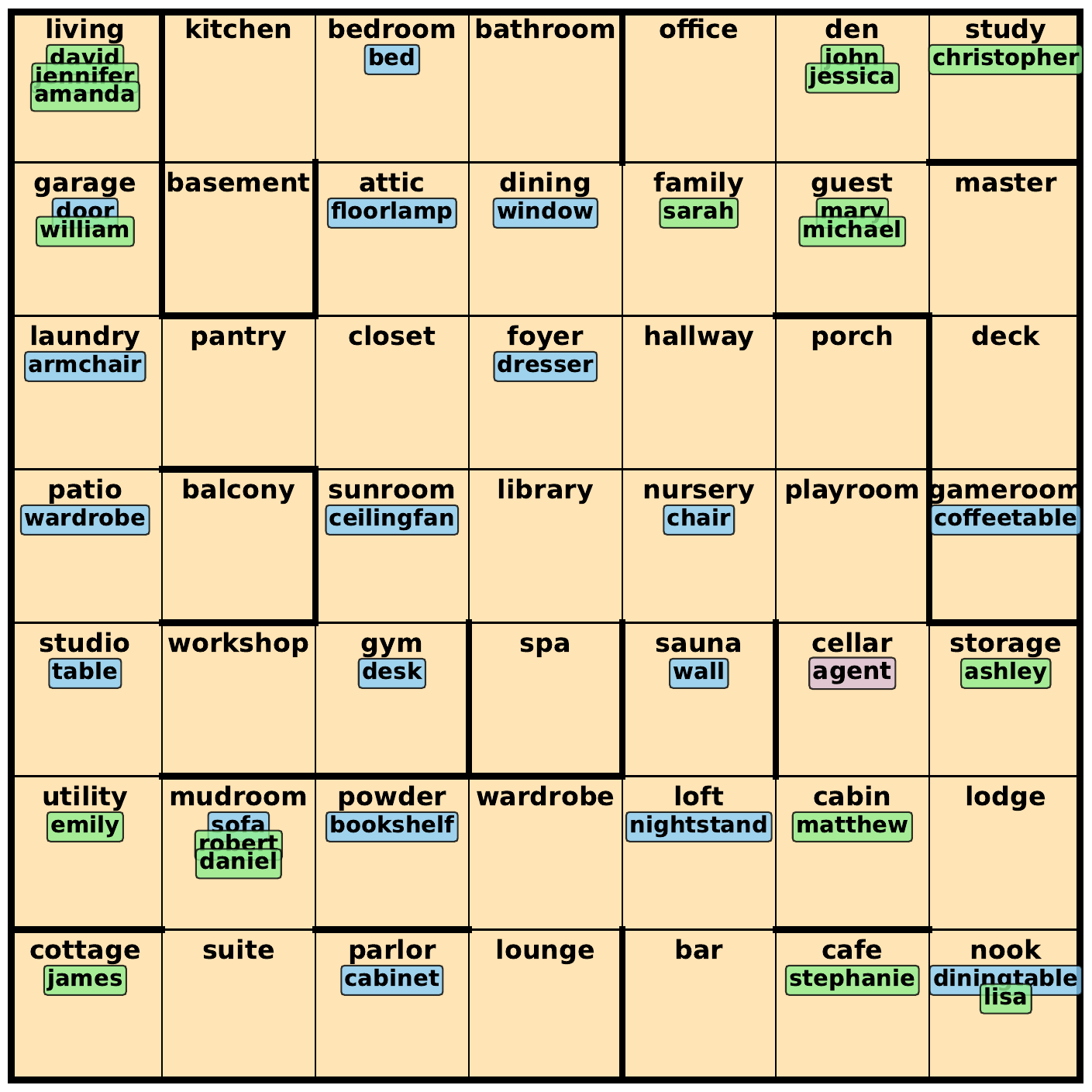}
\caption{Bird's-eye schematic of the hidden state at $t=99$ ($s_{t=99}$) in RoomKG,
showing spatial layout and entity placement. This view is provided for intuition only:
the actual environment state and agent-facing world in our method are represented as RDF
knowledge-graph structures (Figure~\ref{fig:hidden-state-step-099}). The agent does not
directly observe $s_t$; instead, its observation $o_t$ corresponds to a local induced
RDF subgraph around the current room and visible adjacency relations.}
\label{fig:bird-eye-view-step99}
\end{figure}

\begin{figure}[tb]
\centering
\begin{subfigure}[t]{0.55\linewidth}
	\centering
	\includegraphics[width=\linewidth]{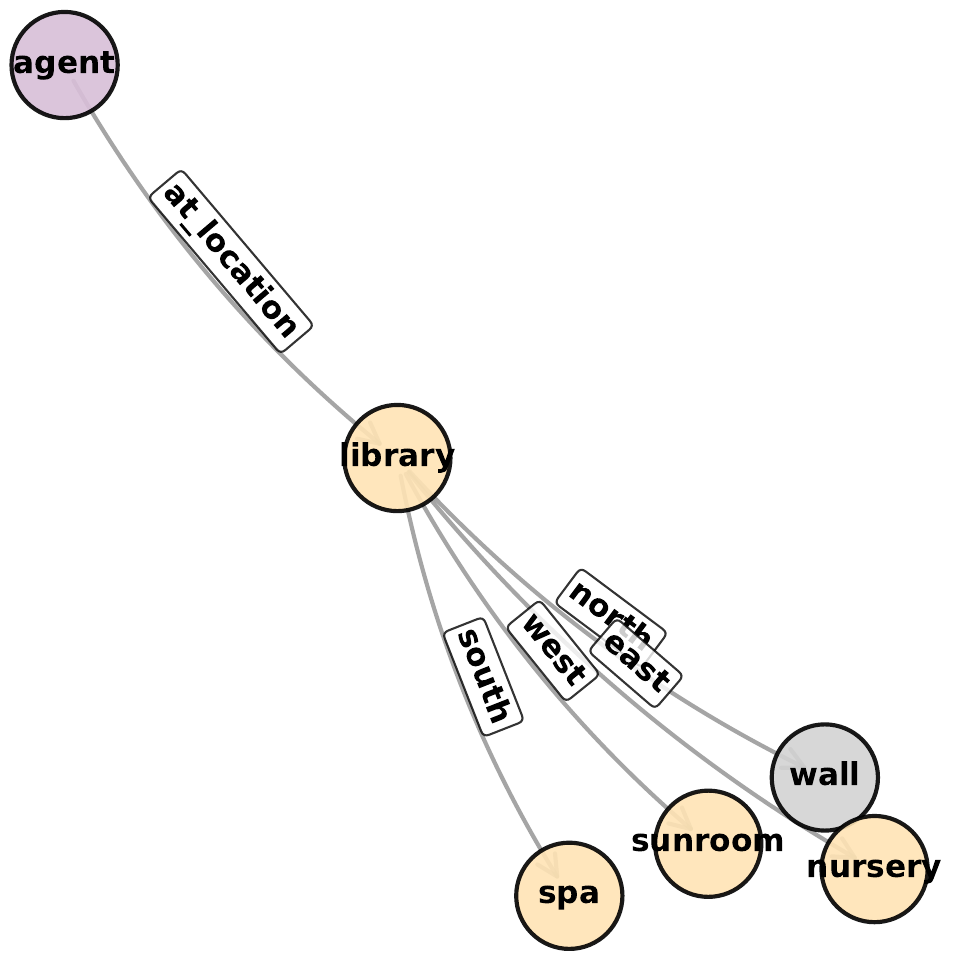}
	\caption{Step 0}
\end{subfigure}

\vspace{0.3em}

\begin{subfigure}[t]{0.82\linewidth}
	\centering
	\includegraphics[width=\linewidth]{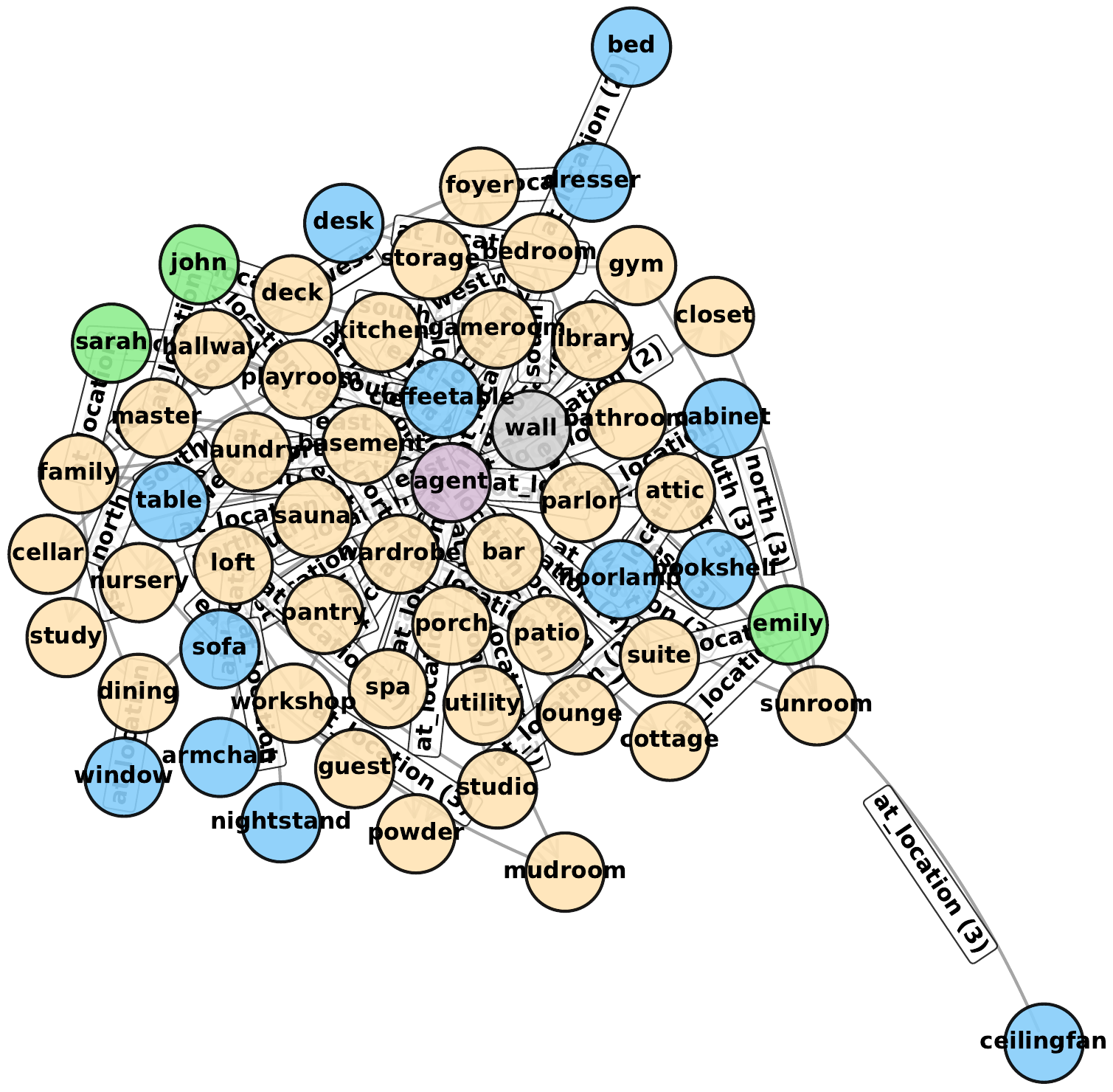}
	\caption{Step 50}
\end{subfigure}
\caption{Agent internal memory-state snapshots at two time points (steps 0 and 50) in a held-out test episode.}
\label{fig:appendix-memory-snapshots-0-50}
\end{figure}

For spatial intuition at the same late-episode point used in the main text,
Figure~\ref{fig:bird-eye-view-step99} shows a bird's-eye schematic of the hidden state
at step $t=99$. Figure~\ref{fig:appendix-memory-snapshots-0-50} complements this view
with two internal memory-state snapshots from held-out test execution (steps 0 and 50).

\end{document}